\documentclass[conference]{IEEEtran}
\IEEEoverridecommandlockouts
\usepackage{cite}
\usepackage{amsmath,amssymb,amsfonts}
\usepackage{algorithmic}
\usepackage{graphicx}
\usepackage{textcomp}
\usepackage{xcolor}
\usepackage{subcaption}
\def\BibTeX{{\rm B\kern-.05em{\sc i\kern-.025em b}\kern-.08em
    T\kern-.1667em\lower.7ex\hbox{E}\kern-.125emX}}
\begin{document}

\title{Data-driven Localization and Estimation of Disturbance in the Interconnected Power System
\thanks{This work was supported by DTRA grants HDTRA1-13-1-0021 and HDTRA1-14-1-0058. The work of Hyang-Won Lee was supported by the National Research Foundation of Korea (NRF) grant funded by the Korea government (MSIT) (No.2018R1D1A1B07048388).}
}

\author{\IEEEauthorblockN{Hyang-Won Lee}
\IEEEauthorblockA{\textit{Dept. of Software},
\textit{Konkuk University}\\
Seoul, Republic of Korea \\
leehw@konkuk.ac.kr}
\and
\IEEEauthorblockN{Jianan Zhang}
\IEEEauthorblockA{\textit{LIDS},
\textit{MIT}\\
Cambridge, MA, USA \\
jianan@mit.edu}
\and
\IEEEauthorblockN{Eytan Modiano}
\IEEEauthorblockA{\textit{LIDS},
\textit{MIT}\\
Cambridge, MA, USA \\
modiano@mit.edu}
}

\maketitle

\begin{abstract}
Identifying the location of a disturbance and its magnitude is an important component for stable operation of power systems. We study the problem of localizing and estimating a disturbance in the interconnected power system. We take a model-free approach to this problem by using frequency data from generators. Specifically,
we develop a logistic regression based method for localization and a linear regression based method for estimation of the magnitude of disturbance. Our model-free approach does not require the knowledge of system parameters such as inertia constants and topology, and is shown to achieve highly accurate localization and estimation performance even in the presence of measurement noise and missing data.
\end{abstract}


\section{Introduction}
Frequency response is one of the key performance measures that indicate the stability of a power system. The frequency of a power system is a complex function of physics, generation control actions and load behaviors over the system topology. Although power systems are designed to operate at a nominal frequency, which is typically 60Hz or 50Hz, they often experience frequency excursion due to the imbalance between generation and load. That is, the system frequency goes up if generation exceeds load, and goes down otherwise. Most of frequency excursions (due to the time-varying nature of demands which generation can keep up with) are not  considered harmful and thus do not call for any action to restore the system frequency. Such a frequency range is referred to as deadband \cite{wecc:tutorial}.

On the other hand, when there is a major disturbance such as generator tripping and load surge,
the frequency can go down to a critical point where generators are damaged permanently or loads severely malfunction. Such a frequency decline should be arrested through so-called primary control that adjusts generation to match the current load, in which case the system frequency reaches the steady state. However, for most of energy sources, it is hard to ramp up generation immediately, and as a consequence, the frequency can decline below a critical point, e.g., 5\% of nominal frequency  \cite{Anderson:adaptive}.

In this case, some loads have to be disconnected from the power grid, and this is referred to as load shedding. Obviously, the amount of load shedding should be neither too small, which may fail to arrest frequency decline below a critical point, nor too large, which may excessively deteriorate the quality of service. It is therefore important to determine the right amount of load to be shed in order to prevent the system frequency from dropping to a critical point. This clearly requires a quick and accurate estimation of the power imbalance between load and generation. In this paper, we study the problem of locating and estimating the disturbance leading to underfrequency, i.e., the disturbance such as load surge and generator tripping.

There are several works that present the method for estimating the power imbalance in the context of load shedding. The work in \cite{Anderson:adaptive} estimates the power imbalance of an isolated generator by measuring the initial slope of frequency decline, i.e., the rate of change of frequency (ROCOF) right after disturbance. The initial slope of an isolated generator is indeed proportional to the power imbalance, which is thus easy to estimate (if the inertia constant is known). This idea can be extended to the case of multiple generators by adding individual power imbalance, which gives the total load-generation imbalance in the power system \cite{Terzija:adaptive}. Furthermore, the accuracy of estimation can be enhanced by using voltage measurements in addition to ROCOF data \cite{Rudez:analysis, Tang:adaptive}, and load characteristics and system topology \cite{Manson:case}.


Most of the above works rely upon the physical model of the power system. This approach, however, hinges on the accurate knowledge of system parameters. For example, as mentioned above, the disturbance estimation method following the principle in \cite{Anderson:adaptive} uses the inertia constant of a power system which also needs to be estimated. Although there are many known methods for estimating the inertia constant, the estimation is subject to several sources of errors and more importantly, the system-wide inertia constant can vary depending on the system status such as load \cite{Chassin:estimation}. In this work, we are interested in the \emph{model-free approach} that does not explicitly use the system model which involves the parameters depending on the fundamental characteristics of the system.

In contrast to the estimation of power imbalance, the literature of localization of disturbance or fault has seen a number of papers taking model-free approaches, i.e., machine learning techniques for fault localization. In \cite{Chen:artificial}, neural networks are used to locate the fault and estimate the fault resistance  based on current and voltage measurement data. In \cite{Cardoso:application}, neural networks are developed separately for different components such as transformers and buses. Only the status of circuit breakers and relays is used as input, and hence, the localization method is scalable and robust to topology changes. Other techniques such as support vector machine \cite{Salat:accurate,Agrawal:identification} and linear discriminant analysis \cite{Zhang:fault} have been used as well (see  \cite{Ferreira:survey} for more references in this context).

There are also several works that take a model-based approach for the localization of a disturbance \cite{Semerow:disturbance}. Typically, in this approach, phasor measurement units (PMUs) detect the disturbance possibly at different times because PMUs are geographically distributed and the disturbance propagates much slower than the electromagnetic wave. The key idea is to use the disturbance propagation speed in order to compute the distance from the disturbance location to each PMU, and apply triangulation method to locate the disturbance. However, the disturbance propagation speed can vary from 100 to 1000 miles/s depending on the system condition \cite{Mei:clustering}.

Our goal in this paper is to develop a disturbance localization and (magnitude) estimation method in the presence of measurement noise and even missing data.
Our method uses frequency data from interconnected generators. 
The interconnected generators are synchronized so that they operate at the same frequency. However, when there is a disturbance, its impact is perceived by generators at different times. For example, a generator close to the disturbance may experience frequency drop earlier than the one far from the disturbance. Consequently, generators may exhibit different frequency dynamics before they are synchronized eventually. As mentioned above, the disturbance propagation speed is slow, and hence, the frequency changes of generators might show distinguishably different patterns depending on the location of disturbance. 
Based on this observation, we apply a simple logistic regression to frequency change data and demonstrates that the location of a disturbance can be identified with high accuracy. In addition, the rate of frequency change also reflects the magnitude of disturbance, and thus, using the same data as in the localization, we propose a simple linear regression based method for estimating the magnitude of disturbance.

The rest of the paper is organized as follows. In Section \ref{sec:model}, we present the model and the problem of our interest. In Section \ref{sec:method}, we discuss the methods for localization and estimation. In Section \ref{sec:simulation}, we demonstrate the performance of our methods under various environments, and in Section \ref{sec:conclusion}, we conclude the paper.

%
%
%
%





%


\section{Model and Problem Description}\label{sec:model}
We consider a power system where there is a control center that collects the frequency data from generators. Note that such a frequency monitoring network (FNET) already exists, and there is even a low cost 120V-outlet measurement based FNET \cite{Zhang:wide,FNET:website}.
Assume that there are $N$ generators and $B$ buses. Let $f_i(t)$ be the frequency of generator $i$ at time $t$. Suppose that there is a sudden increase of load at a bus. Let $\Delta P_i$ denote the power imbalance, i.e., load minus generation, at generator $i$. This value can be expressed as
\begin{align}
\Delta P_i = -\frac{2H_i S_i}{f_n} \frac{df_{i}(t)}{dt},\label{eqn:dis-freq-single}
\end{align}
where $H_i$[s], $S_i$[MVA] and $f_i$[Hz] are the inertia constant, rated apparent power and frequency of generator $i$, respectively, and $f_n[Hz]$ is the nominal frequency \cite{Anderson:adaptive}. To be more precise, $\Delta P_i$ is given by $\Delta P_i=P_{e_i}-P_{m_i}$ where $P_{m_i}$ and $P_{e_i}$ are mechanical input power to generator and electric output power from generator, respectively. Hence,  \eqref{eqn:dis-freq-single} represents the fact that when the system load suddenly increases, the rotational energy in the mass of generator unit is released to initially supply the load, thereby decreasing the frequency. It is important to note that this relationship between power imbalance and rate of frequency change is valid only right after the disturbance has occurred. This is because once the control action (specifically, primary control that immediately responds to frequency change) of generator takes effect, the effect of disturbance decays and the system reaches the steady state where $\frac{df_{i}(t)}{dt}\approx0$.

In the interconnected power system, a load change is shared by generators. This is typically expressed by summing the individual power imbalances as
\begin{align}
\Delta P = \sum_{i}\Delta P_{i} = -\sum_{i}\frac{2H_i S_i}{f_n} \frac{df_{i}(t)}{dt}, \label{eqn:dis-freq-sum}
\end{align}
where $\Delta P$ is the total power imbalance in the system
 \cite{Tang:adaptive}.
Using this model, the total power imbalance in the system can be estimated using the rate of frequency change from each generator. This estimation, however, can suffer from several errors. First, as mentioned in the introduction, the inertia constant $H_i$ can change depending on the system status such as load. Second, if there is a disturbance at a certain bus, the disturbance starts to take effect at a generator nearest to the bus. The electric output power at the nearest generator will then suddenly increase, leading to frequency decline, while other generators may not have received the impact yet. This makes it unclear when the model in \eqref{eqn:dis-freq-sum} should be used to estimate the disturbance. This subtlety of model-based approach has led us to consider a model-free approach that does not relay on a specific system model.

Our approach uses the rate of change of frequency (ROCOF) as well because it reflects the magnitude of disturbance in that the frequency change is larger in the event of larger disturbance. In this paper, we assume that the disturbance occurs at a single bus (our method can be readily extended to the case of multiple disturbances), and that the disturbance start time is known (there are existing methods for detecting start time such as the one in \cite{Li:online}). We first develop a localization method by applying logistic regression to ROCOF data. The location information together with ROCOF data is used to estimate the magnitude of disturbance based on linear regression.

%


\section{Perturbation Localization and Magnitude Estimation}\label{sec:method}
In this section, we present the method of disturbance localization based on logistic regression, and disturbance magnitude estimation based on linear regression. We first discuss the selection of features for the learning algorithms, using the frequency measurement at all the generators. Furthermore, we discuss how training and prediction are performed when some measurements are missing, e.g., due to communication delays or failures.

\subsection{Extracting Features from Frequency Data}
Assume that the frequency at each generator is measured with a PMU. Typically, PMUs are equipped with GPS for clock synchronization, and hence, measurement data can be assumed synchronized. Let $t_k$ be the $k$-th sample time at every generator. The control center receives from each generator $i$ the noisy frequency measurement expressed as $\tilde{f}_i(t_k) = f_i(t_k) + \epsilon_i$
where $\epsilon_i$ is the measurement noise normally distributed with mean zero and variance $\sigma^2$, i.e., $\epsilon_i\sim\mathcal{N}(0,\sigma^2)$. Real frequency measurement data show that measurement noise is common and non-negligible \cite{Li:online}.


Recall that the time at which the disturbance occurred is assumed to be known. Without loss of generality, let $t_0$ be the time at which the disturbance occurred. As discussed in Section \ref{sec:model}, the initial slope $f_i'(t_0)\triangleq\frac{f_i(t_1)-f_i(t_0)}{t_1-t_0}$ is proportional to the magnitude of the disturbance. 
Depending on the distance from the epicenter, the initial slopes at a generator may exhibit different patterns for different locations of disturbance.
One could use more samples afterwards, $f_i'(t_1),f_i'(t_2),...$, so as to construct a more distinguishable footprint of disturbance.

Note that when there is a disturbance, the frequency declines and thus the slope is negative. However, with actual noisy frequency data, the sign of slope $\tilde{f}_i'(t_k)\triangleq\frac{\tilde{f}_i(t_{k+1})-\tilde{f}_i(t_k)}{t_{k+1}-t_k},k=0,1,2,...$ can fluctuate, even if the frequency data are smoothed, whereas the original signs are steadily negative. This is detrimental to training, as the sign (and magnitude as well) of the slope captures the critical information of disturbance.

To address this issue, we use the following form of frequency change. Let $\Delta\tilde{f}_i(t_k)=\tilde{f}_i(t_k)-\tilde{f}_i(t_0),k=1,2,...$ for each generator $i$. Assuming equally spaced sample times, the values represent the slopes with respect to the disturbance moment $t_0$. Clearly, compared to the values $\tilde{f}_i'(t_k)$ defined above, the signs of $\Delta\tilde{f}_i(t_k)$ are more likely to be the same as the original signs of $\Delta f_i(t_k)$ with noiseless data because it considers difference between (originally declining) frequency values farther separated in time. Hence, this gives a more robust measure of frequency change. We apply a simple mean filter to these values as
\begin{align}
x_i =\begin{bmatrix}
\frac{1}{W_a} \sum\limits_{k=j}^{j+W_a-1} \Delta\tilde{f}_i(t_k),\,\, j=1,...,W_s-W_a+1
\end{bmatrix},
\end{align}
and we use the following vector $x$ as a feature vector:
\begin{align}
x = \begin{bmatrix}
x_1 & x_2 & \cdots & x_N & 1\label{eqn:feature-vec}
\end{bmatrix}.
\end{align}
The $k$th coordinate of $x_i$ will be denoted as $x_{ik}$. Here, the value $W_a$ is the \emph{averaging window size} that determines the smoothness of the filtered data in $x_i$, i.e., large values of $W_a$ lead to smoother data. The value $W_s$ is the \emph{sampling window size} that determines how many samples will be used to form a feature vector. 
The last scaler value 1 is used to fit the intercept.
The length of the feature vector is $L=(W_s - W_a + 1)N + 1$, where $N$ is the number of PMUs.
In Section \ref{sec:simulation}, we examine the impact of these values on the accuracy of prediction.

%
%


\subsection{Logistic Regression for Localization}
Let $y=[y_0\,\, y_1\,\,...\,\,y_B]$ be a binary vector such that $y_i=1$ if there is a disturbance at bus $i$, and $y_i=0$ otherwise. The value $y_0=1$ indicates that there is no disturbance. Since we assume that there is at most one disturbance, we have $\sum_{i=0}^By_i=1$. One sample of the training data for localization is given by $(y,x)$, i.e., $x$ is the frequency change information in \eqref{eqn:feature-vec} and $y$ is the true disturbance location. Denote by $X$ the random variable representing the feature vector $x$, and $Y$ the indicator vector $y$. The randomness comes from the measurement noise. The logistic regression problem is formulated as
\begin{align}
\max_{\beta=[\beta^0\,\,\beta^1\,\,\cdots\,\,\beta^B]} \mathbb{P}(Y|X) = \prod_{b=0}^B p_b(X;\beta)^{Y_b},\label{eqn:logreg}
\end{align}
where $p_b(X)$ represents the probability that the disturbance occurs at bus $b$. Typically, this probability is expressed as
\begin{align}
p_b(X;\beta) = \frac{e^{\beta^b\cdot X}}{\sum_{a=0}^B e^{\beta^a\cdot X}},
\end{align}
where $\beta^b\in\mathbb{R}^{L}$ is the coefficient vector corresponding to the disturbance at bus $b$, and
\begin{align}
\beta^b\cdot X = \beta^b_0 + \sum_{i=1}^{N}\sum_{k=1}^{W_s-W_a+1} \beta_{ik}^b X_{ik}.
\end{align}

Suppose that there are $M$ samples of training data, $(y^1,x^1),...,(y^M,x^M)$. Taking the logarithmic function on the objective function in \eqref{eqn:logreg} and adding a regularization term, the logistic regression problem computes $\beta$ that minimizes the objective function
\begin{align}
\min_\beta \frac{\lambda}{2}||\beta||_2^2 - \sum_{j=1}^M\left\{\sum_{b=0}^B y_b^j(\beta^b\cdot x^j) - \log\left(\sum_{b=0}^B e^{\beta^b\cdot x^j}\right)\right\},\label{eqn:logreg-data}
\end{align}
where $\lambda$ is a parameter that controls the strength of regularization. For large values of $\lambda$, regularization is emphasized so as to avoid over-fitting and enhance robustness to noise.

Denote by $\hat{\beta}$ the solution of \eqref{eqn:logreg-data}. Given a new measurement data $x^{\mbox{new}}$, the location of a disturbance is estimated as the bus $b$ with the largest $p_b(x^{\mbox{new}};\hat{\beta})$. The advantage of logistic regression is that the value $p_b(x^{\mbox{new}};\hat{\beta})$ gives the probability that $b$ is the location of disturbance, and hence, these values can be used to pick, say $k$, most probable locations of disturbance. These candidates can be further examined by other methods or even human experts in order to narrow down to the actual location. In Section \ref{sec:simulation}, we show that this method can significantly reduce the classification error.

\subsection{Estimation of Magnitude of Disturbance}

Given the location of a disturbance, we now discuss how the magnitude of the disturbance is estimated. As discussed with respect to the relationship \eqref{eqn:dis-freq-single}, the magnitude of initial frequency change increases as the magnitude of disturbance increases. We thus use the same feature $x$ as in the localization above, which gives a robust measure of frequency change after a disturbance has occurred. Furthermore, as the model \eqref{eqn:dis-freq-sum} suggests, one can expect a linear relationship between rate of frequency change and magnitude of disturbance. This led us to apply linear regression as it seeks to find an affine function that best fits the input ($x$ in our case) and output ($\Delta P$ in our case) data.

Unlike in the localization, we train the linear regression separately for each disturbance location, i.e., separate linear regression for each bus. Consider an arbitrary bus $b$. Let $z\in\mathbb{R}_{+}$ be the magnitude of a disturbance that has occurred at bus $b$. One sample of training data is then given as $(z,x)$. Assuming that there are $M$ samples $(z^1,x^1),...,(z^M,x^M)$, the linear regression problem is formulated as
\begin{align}
\min_{\alpha^b} \frac{1}{2}\sum_{j=1}^M\left(z^j-\alpha^b\cdot x^j \right)^2.\label{eqn:linreg-data}
\end{align}
Let $A=[ x^1;\,\,\cdots;\,\,x^M]\in\mathbb{R}^{M\times L}$ and $q=[z^1;\,\,\cdots\,\,;z^M]\in\mathbb{R}^M$. As long as $A$ has full column rank, the solution to the above minimization problem is given as $\hat{\alpha}^b = (A^TA)^{-1}A^T q$. 
Otherwise, if the inverse is computationally expensive or does not exist, the solution can be found using gradient descent methods.
Given the new measurement data $x^{\mbox{new}}$ and location $b$ (either from the localization above or known a priori), the magnitude is estimated as $\hat{\alpha}^b\cdot x^{\mbox{new}}$.

\subsection{Dealing with Missing Data}
So far, we have assumed that the communication between control center and PMUs is always reliable. However, as PMUs are increasingly deployed over power grid, the communication channel can possibly turn unreliable due to excessive amount of data or even link failures. It is therefore important to ensure the above regression schemes work as designed even when some data from generators are missing at the time of prediction. 

Assume for simplicity that measurement data is missing from at most one generator (we will discuss the general case of missing data below). To address the scenario of missing data, for each missing scenario, we train logistic and linear regression coefficients in advance. This requires $N+1$ separate trainings for both localization and estimation since we assume missing data from at most one generator. Note that everything in the training process is identical to the case of no missing data except that the feature vector $x$ in \eqref{eqn:feature-vec} does not include the measurement at the generator where the data are missing.
To address the general case where data can be missing from at most $k$ generators, one can separately train for $\sum_{j=0}^{k}\binom{N}{k}$ missing scenarios. Similarly, as soon as the control center finds that measurement data are missing from $j$ generators $\{G_{i_1},...,G_{i_j}\}$, the regression coefficients corresponding to missing generators $\{G_{i_1},...,G_{i_j}\}$ are retrieved and used for localization and estimation. We believe that this method is a practical solution because the event that a large number of generators fail to deliver their data at the same time may be unlikely. In the next section, we show that even with missing data, our schemes yield fairly accurate localization and estimation performance.

\section{Simulation}\label{sec:simulation}
We generate the data using MATLAB power system toolbox (Simscape Power Systems), based on the New England Power System IEEE benchmark topology \cite{moeini2015open}. Each disturbance scenario is a load increase at one of 21 bus locations. For the training data, the load increases range from 100 MW to 1000 MW, with 10 MW interval. For the test and validation data, the load increases are chosen uniformly at random within the range from 100 MW to 1000 MW. The frequencies are sampled at all generators once every 5 millisecond.

In the rest of this section, we study the performance of the learning algorithms on disturbance localization and magnitude predictions, using scikit-learn package \cite{scikit-learn}. The hyper-parameters of the learning algorithms, such as the regularization coefficient, are tuned to minimize the prediction error in the test data. We then evaluate the performance of the algorithms using validation data, which avoids over-fitting of the algorithms on test data.

\subsection{Disturbance localization}
We evaluate the performance of logistic regression in disturbance localization. We first explain the tuning of regularization coefficient under noisy measurement, and then evaluate the prediction accuracy by tuning the sampling window size $W_s$ and averaging window size $W_a$. We report the optimal $W_s$ and $W_a$ that minimize the error rates, and provide intuitions on the selections of $W_s$ and $W_a$.   


\subsubsection{Regularization coefficient $\lambda$}
Recall that $\lambda$ is tuned to minimize the test error. Table \ref{tb:noise} shows the error rates (defined as the fraction of misclassified scenarios) on test and validation data under $W_s = 200, W_a = 1$, and optimal $\lambda$. We observe that, as noise magnitude increases, a larger regularization strength is required, and the estimation error increases. Since the error rates on test data and validation data are close, in the remainder of the section, we only report the error rates on validation data to study the impact of $W_s$ and $W_a$ under the optimally tuned regularization coefficient.

\begin{table}[h]
\centering
\caption{Classification errors under different noise levels}
\label{tb:noise}
\vspace{-0.15cm}
\begin{tabular}{|l|l|l|l|l|l|}
\hline
$\sigma$ (mHz)    & 0     & 0.5   & 1      & 5      & 10      \\ \hline
$\lambda$                & 1     & 100  & $2\times 10^3$ & $10^4$ & $10^5$ \\ \hline
test error       & 0.006 & 0.015 & 0.033  & 0.103  & 0.149   \\ \hline
validation error & 0.009 & 0.012 & 0.030  & 0.118  & 0.155   \\ \hline
\end{tabular}
\end{table}

\subsubsection{Sampling window size $W_s$}
First, we study the prediction accuracy using noiseless frequency measurement data. For sampling window size $W_s = 1$, the error rate on validation data is around $0.9\%$. For $W_s = 2$, the error rate is zero for the validation data. A small number of frequency samples after the disturbance are sufficient to achieve high prediction accuracy. This demonstrates the effectiveness of logistic regression in locating the disturbance location. 

Then, we study the predictions under noisy measurement. We assume that frequency measurement errors follow a Gaussian distribution with standard deviation $\sigma = 5$mHz \cite{Li:online}, and that the measurement errors are independent. From Table \ref{tb:m}, we observe that the classification errors are very high for small $W_s$. A larger $W_s$ is required to achieve higher accuracy, and $W_s = 200$ achieves the highest accuracy among our tests.

\begin{table}[h]
\centering
\caption{Classification errors under different sampling window sizes}
\label{tb:m}
\vspace{-0.15cm}
\begin{tabular}{|l|l|l|l|l|l|}
\hline
$W_s$              & 5     & 50    & 100  & 200 & 500 \\ \hline
validation error & 0.709 & 0.185 & 0.140 &0.118 &0.145
 \\ \hline
\end{tabular}
\end{table}

\begin{figure}[h]
  \begin{subfigure}[b]{0.24\textwidth}
    \includegraphics[width=\textwidth]{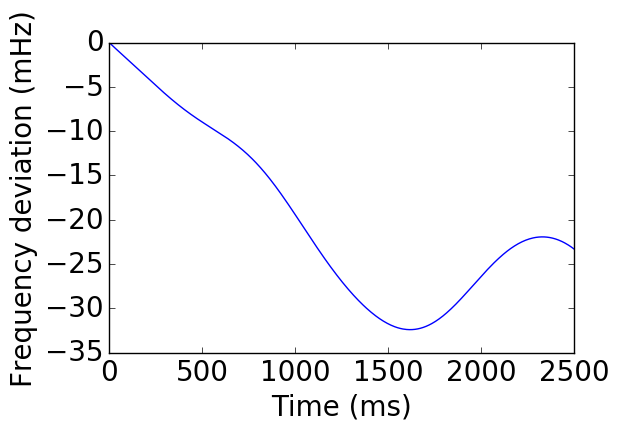}
    \caption{Generator 1, without noise.}
    \label{fig:1}
  \end{subfigure}
  \begin{subfigure}[b]{0.24\textwidth}
    \includegraphics[width=\textwidth]{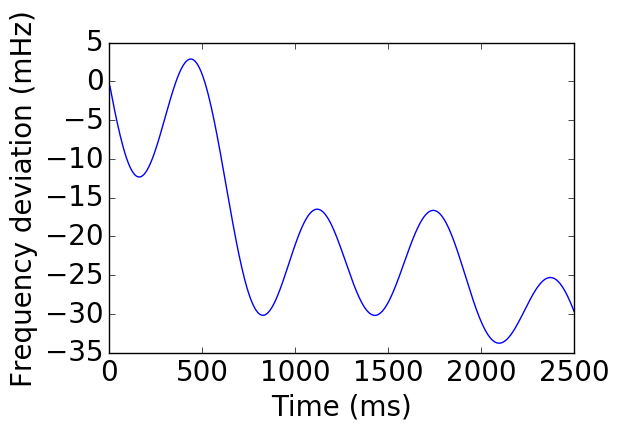}
    \caption{Generator 10, without noise.}
    \label{fig:2}
  \end{subfigure}
  \begin{subfigure}[b]{0.24\textwidth}
    \includegraphics[width=\textwidth]{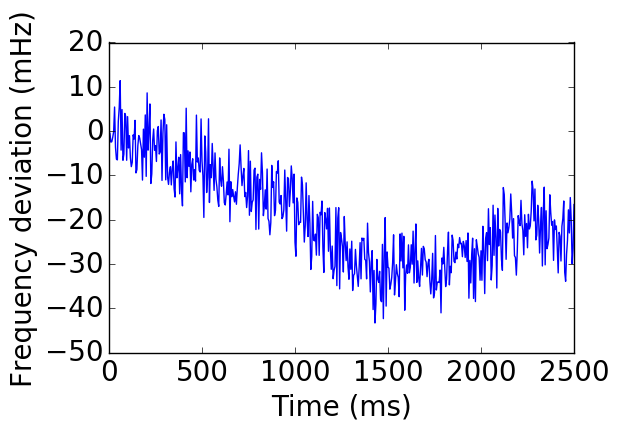}
    \caption{Generator 1, with noise.}
    \label{fig:3}
  \end{subfigure}
  \begin{subfigure}[b]{0.24\textwidth}
    \includegraphics[width=\textwidth]{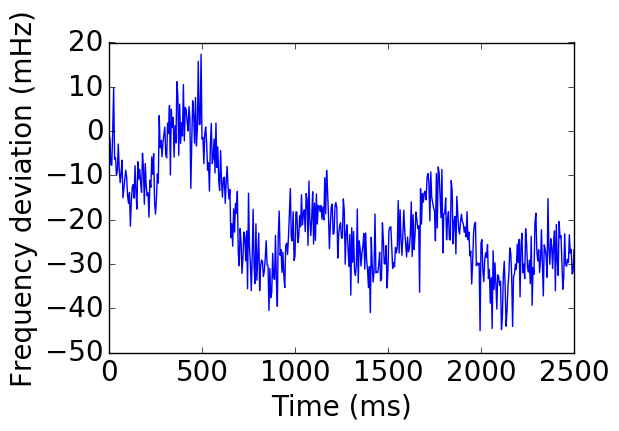}
    \caption{Generator 10, with noise.}
    \label{fig:4}
  \end{subfigure}
  \vspace{-0.5cm}
  \caption{Frequency deviation measurements with and without noise ($\sigma = 5$mHz).}
  \vspace{-0.0cm}
  \label{fig:conv}
\end{figure}

Figure \ref{fig:conv} illustrates the frequency deviations within 2500ms ($W_s = 500$) at generators 1 and 10, after 200MW load increase at bus 4. Since the frequency changes are non-linear and the generators respond quickly to power imbalance, the rate of change within a small time interval after the disturbance provides accurate estimation on the disturbance location. Therefore, $W_s = 2$ gives zero error when there is no measurement noise. However, with measurement noise, the rate of change cannot be recovered from a small $W_s$. Instead, the rate can only be estimated when $W_s$ is sufficiently large, so that the frequency deviation becomes large relative to the noise. On the other hand, a larger $W_s$ increases both the computation cost and the difficulty to regularize the learning algorithm to avoid over-fitting. Therefore, a moderate value of $W_s$ is required to minimize the estimation error.

\subsubsection{Averaging window size $W_a$}
We study the effects of smoothing on reducing the estimation error. Table \ref{tb:n} shows the error rates for different averaging window sizes, when $\sigma = 5$mHz and $W_s = 200$. We observe that smoothing reduces the classification errors. As $W_a$ increases, the number of features ($W_s - W_a + 1$) associated with each generator decreases, and the training algorithm runs faster. We observe that the error rates are low for $W_a = [50,100]$.

\begin{table}[h]
\centering
\caption{Classification errors under different averaging window sizes}
\label{tb:n}
\vspace{-0.15cm}
\begin{tabular}{|l|l|l|l|l|l|}
\hline
$W_a$               & 1     & 10  & 50   & 100   & 150   \\ \hline
validation error & 0.118 & 0.100 & 0.060 & 0.057 & 0.072 \\ \hline
\end{tabular}
\end{table}

\subsubsection{Top $k$ most likely disturbance locations}
For the mis-classified scenarios, logistic regression still provides useful information for disturbance location. In addition to the predicted disturbance location, logistic regression outputs the probabilities of each disturbance location. The top $k$ locations that have higher probabilities form a set of locations that require further inspection. Table \ref{tb:set} shows the fraction of validation data where the top $k$ scenarios fail to include the true disturbance location, for $W_s = 200, W_a = 100,\sigma=5$mHz. We observe that, among $2/3$ of the mis-classified scenarios, the top two most likely locations estimated by the logistic regression include the true disturbance location.
\begin{table}[h]
\centering
\caption{Fraction of data where the top $k$ estimated locations fail to contain the true location}
\label{tb:set}
\vspace{-0.15cm}
\begin{tabular}{|l|l|l|l|l|l|}
\hline
$k$                & 1     & 2     & 3     & 4     & 5     \\ \hline
validation error & 0.057 & 0.021 & 0.015 & 0.012 & 0.009 \\ \hline
\end{tabular}
\end{table}

\subsubsection{Missing measurement}
Finally, we evaluate the prediction accuracy when there are missing measurements from one or more generators. Table \ref{tb:missing} shows the average classification errors when the frequency measurement from $i$ generators are missing, $i \in \{1,2,3,4,5\}$, for $W_s=200, W_a=100,\sigma=5$mHz. We observe that the predictions are robust under single generator measurement failure, and that the error rate increases as there are more number of missing measurements.

\begin{table}[h]
\centering
\caption{Classification errors under missing measurement}
\label{tb:missing}
\vspace{-0.15cm}
\begin{tabular}{|l|l|l|l|l|l|}
\hline
$i$ & 1     & 2     & 3    & 4     & 5     \\ \hline
validation error  & 0.066 & 0.082 & 0.090 & 0.116 & 0.137 \\ \hline
\end{tabular}
\end{table}

These results show that by tuning the window size parameters, our localization method can provide a robust classification in the face noise and unreliable communications.

\subsection{Disturbance magnitude estimation}
We apply linear regression to estimate the disturbance magnitude, given disturbance location. We observe that both $l_1$ and $l_2$ regularizations do not improve the prediction accuracy. Therefore, we apply ordinary least square estimation \eqref{eqn:linreg-data}, and report the average relative errors $|(\Delta \hat{P}-\Delta P)/\Delta P|$ on the validation data.

Table \ref{tb:noise_reg} shows the relative errors under different noise levels, for $W_s = 200, W_a = 1$. We observe that the errors are negligible for moderate measurement noise. Table \ref{tb:sampling} shows that the error decreases as the sampling window size $W_s$ increases. The averaging window size does not have noticeable impact on the errors, and the numerical results are omitted.
\begin{table}[h]
\centering
\caption{Regression errors under different noise levels}
\label{tb:noise_reg}
\vspace{-0.15cm}
\begin{tabular}{|l|l|l|l|l|l|}
\hline
$\sigma$ (mHz)         & 0        & 1      & 5     & 10    \\ \hline
relative error & $8.4\times10^{-5}$ & $8\times10^{-4}$ & $3\times10^{-3}$ & $7\times10^{-3}$ \\ \hline
\end{tabular}
\end{table}

\begin{table}[h]
\centering
\caption{Regression errors under different sampling window sizes ($\sigma=5$mHz)}
\label{tb:sampling}
\vspace{-0.15cm}
\begin{tabular}{|l|l|l|l|l|l|}
\hline
$W_s$                & 5     & 50    & 100   & 200   & 500   \\ \hline
relative error & 0.411 & 0.028 & 0.016 & 0.007 & 0.003 \\ \hline
\end{tabular}
\end{table}

\begin{figure}[h]
\centering
\includegraphics[width=.8\linewidth]{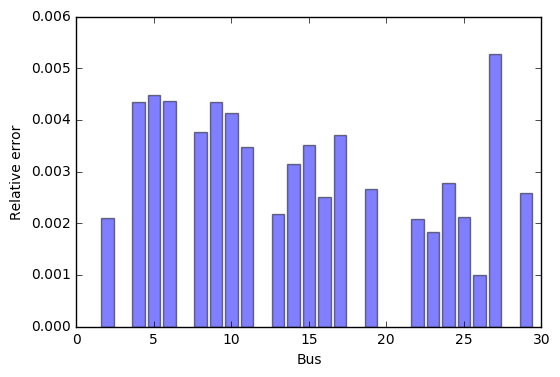}
\vspace{-0.2cm}
\caption{Magnitude estimation error at each bus location.}
\vspace{-0.0cm}
\label{fig:comp}
\end{figure}

The estimation accuracy differs at different disturbance locations. Figure \ref{fig:comp} illustrates the average relative errors for disturbances at each bus locations, where the $x$-axis represents the bus ID, and the $y$-axis represents the average relative error. We observe that estimation errors are generally smaller for disturbances at bus locations close to generators (i.e., PMU locations), such as buses 2, 23, 25, 26.

The disturbance magnitude estimations are robust under missing measurement. Remarkably, when there is available measurement at only one generator, the relative error only increases to 0.020, for any given disturbance location. This can be explained by the observation that the amount of frequency deviation is a monotone function in the amount of disturbance, for a fixed disturbance location.

We remark that the estimation of disturbance magnitude based on Eq. (\ref{eqn:dis-freq-sum}) has very high error (over $60\%$ relative error), even without measurement noise. This further demonstrates the superior performance of our estimation algorithms.

\section{Concluding Remarks}\label{sec:conclusion}
We developed logistic regression and linear regression based methods for localizing and estimating a disturbance by using frequency data from generators. Our model-free approach does not require the knowledge of system parameters such as inertia constants and topology. We showed through simulations that our approach achieves highly accurate localization and estimation of a disturbance even in the presence of measurement noise and missing data. The power system increasingly integrates distributed generation and renewable resources that bring about more uncertainty compared to the typical large-scale power plant. The traditional model-based approach to the localization and estimation problem hinges on the accuracy of a model, which may be highly challenging with increasing uncertainty. Our results in this paper serves as an example that shows the effectiveness of model-free approaches applying machine learning techniques in the face of increasingly complex power system.





\bibliographystyle{IEEEtran}
\bibliography{reference}

\end{document}